\documentclass[runningheads]{llncs}
\usepackage{subcaption}
\usepackage{graphicx}
\usepackage{amsmath,amssymb} 
\usepackage{commath}
\usepackage{color}
\usepackage{cite}
\usepackage{epstopdf}
\usepackage{float}
\usepackage{array}
\usepackage{enumerate}
\usepackage[width=122mm,left=12mm,paperwidth=146mm,height=193mm,top=12mm,paperheight=217mm]{geometry}
\usepackage{hyperref}
\DeclareMathOperator*{\argmin}{argmin}
\begin{document}
\pagestyle{headings}
\mainmatter

\title{Two-stage Convolutional Part Heatmap Regression for the 1st 3D Face Alignment in the Wild (3DFAW) Challenge} 

\titlerunning{Two-stage Convolutional Part Heatmap Regression for the 1st 3D Face Alignment in the Wild (3DFAW) Challenge}

\authorrunning{Adrian Bulat \and Georgios Tzimiropoulos}

\author{
	Adrian Bulat 
	\and 
	Georgios Tzimiropoulos
}


\institute{Computer Vision Laboratory, University of Nottingham, Nottingham, UK
	\email{\{adrian.bulat,yorgos.tzimiropoulos\}@nottingham.ac.uk}
}

\maketitle

\begin{abstract}
This paper describes our submission to the 1st 3D Face Alignment in the Wild (3DFAW) Challenge. Our method builds upon the idea of convolutional part heatmap regression \cite{myECCV2016}, extending it for 3D face alignment. Our method decomposes the problem into two parts: (a) X,Y (2D) estimation and (b) Z (depth) estimation. At the first stage, our method estimates the X,Y coordinates of the facial landmarks by producing a set of 2D heatmaps, one for each landmark, using convolutional part heatmap regression. Then, these heatmaps, alongside the input RGB image, are used as input to a very deep subnetwork trained via residual learning for regressing the Z coordinate. Our method ranked 1st in the 3DFAW Challenge, surpassing the second best result by more than 22\%. Code can be found at \url{http://www.cs.nott.ac.uk/~psxab5/}
\end{abstract}

\section{Introduction}

Face alignment is the problem of localizing a set of facial landmarks in 2D images. It is a well-studied problem in Computer Vision research, yet most of prior work \cite{Cao2012shaperegression, xiong2013supervised}, datasets \cite{sagonas2013semi, sagonas2013300} and challenges  \cite{sagonas2013300,Shen15} have focused on frontal images. However, under a totally unconstrained scenario, faces might be in arbitrary poses. To address this limitation of prior work, recently, a few methods have been proposed for large pose face alignment \cite{jourabloo2015pose,jourabloo2016large,zhu2015face,myBMVC2016}. 3D face alignment goes one step further by treating the face as a full 3D object and attempting to localize the facial landmarks in 3D space. To boost research in 3D face alignment, the 1st Workshop on 3D Face Alignment in the Wild (3DFAW) \& Challenge is organized in conjunction with ECCV 2016 \cite{Workshop3DFAW}. In this paper, we describe a Convolutional Neural Network (CNN) architecture for 3D face alignment, that ranked 1st in the 3DFAW Challenge, surpassing the second best result by more than 22\%.

Our method is a CNN cascade consisting of three connected subnetworks, all learned via residual learning \cite{he2015deep,he2016identity}. See Fig. \ref{fig:Our3DNET}. The first two subnetworks perform residual part heatmap regression \cite{myECCV2016} for estimating the X,Y coordinates of the facial landmarks. As in \cite{myECCV2016}, the first subnetwork is a part detection network trained to detect the individual facial landmarks using a per-pixel softmax loss. The output of this subnetwork is a set of N landmark detection heatmaps. The second subnetwork is a regression subnetwork that jointly regresses the landmark detection heatmaps stacked with image features to confidence maps representing the location of the landmarks. Then, on top of the first two subnetworks, we added a third very deep subnetwork that estimates the Z coordinate of each fiducial point. The newly introduced network is guided by the heatmaps produced by the 2D regression subnetwork and subsequently learns where to ``look'' by explicitly exploiting information about the 2D location of the landmarks. We show that the proposed method produces remarkable fitting results for both X,Y and Z coordinates, securing the first place on the 3DFAW Challenge.

This paper is organized as follows: section \ref{sec:method} describes our system in detail. Section \ref{sec:results} describes the experiments performed and our results on the 3DFAW dataset. Finally, section \ref{sec:conclusions} summarizes our contributions and concludes the paper.

\section{Method}\label{sec:method}
\vspace{-2em}
\begin{figure}
	\centering 
    \hspace{-1em}
    \vspace{-1em}
	\includegraphics[scale=0.16]{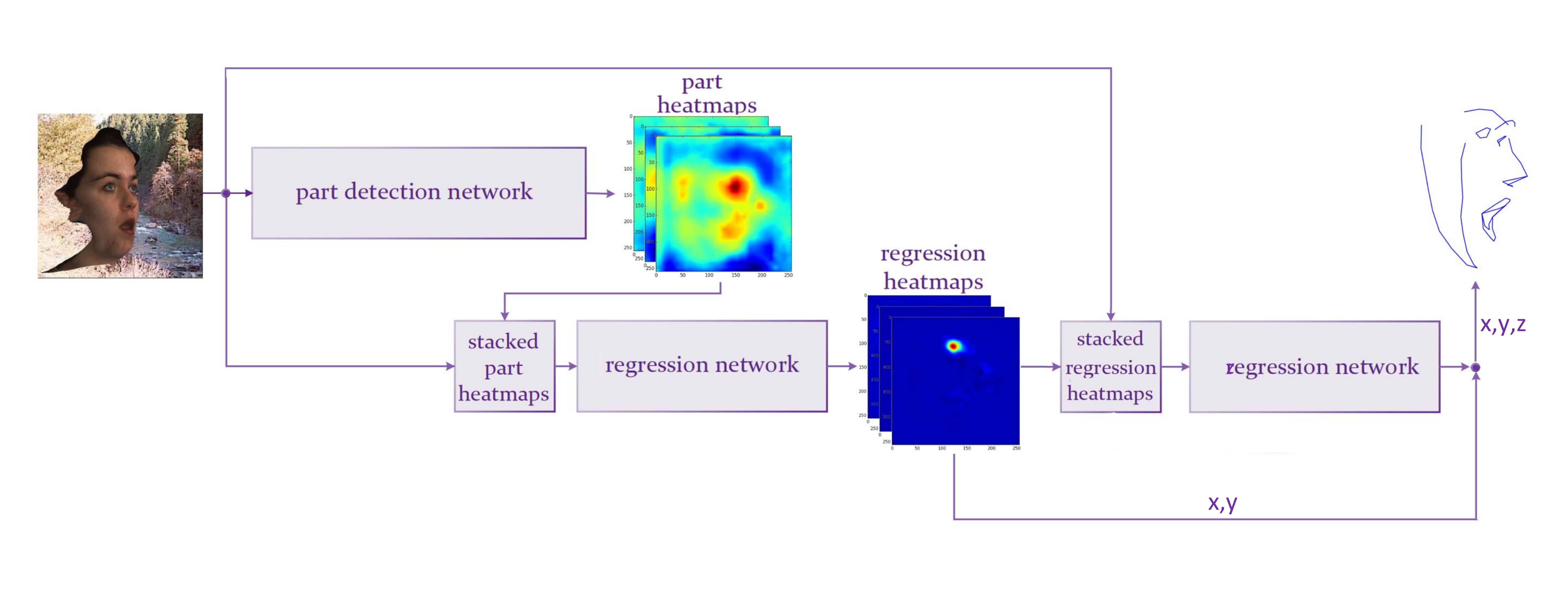}
	\caption{The system submitted to the 3DFAW Challenge. The part detection and regression subnetworks implement convolutional part heatmap regression, as described in \cite{myECCV2016}, and produce a series of N heatmaps, one for the X,Y location of each landmark. They are both very deep networks trained via residual learning \cite{he2016identity}. The produced heatmaps are then stacked alongside the input RGB image, and used as input to the Z regressor which regresses the depth of each point. The architecture for the Z regressor is based on ResNet \cite{he2016identity}, as described in section \ref{ssec:method1d}}.
	\label{fig:Our3DNET}
\end{figure} 

The proposed method adopts a \textit{divide et impera} technique, splitting the 3D facial landmark estimation problem into two tasks as follows: The first task estimates the X,Y coordinates of the facial landmarks and produces a series of N regression heatmaps, one for each landmark, using the network described in section \ref{ssec:method2d}. The second task, described in section \ref{ssec:method1d}, predicts the Z coordinate (i.e. the depth of each landmark), using as input the stacked heatmaps produced by the first task, alongside the input RGB image. The overall architecture is illustrated in \figurename{\ref{fig:Our3DNET}}.

\subsection{2D (X,Y) landmark heatmap regression}\label{ssec:method2d}

The first part of our system estimates the X,Y coordinates of each facial landmark using part heatmap regression \cite{myECCV2016}. The network consists of two connected subnetworks, the first of which performs landmark detection while the second one refines the initial estimation of the landmarks' location via regression. Both networks are very deep been trained via residual learning \cite{he2015deep,he2016identity}. In the following, we briefly describe the two subnetworks; the exact network architecture and layer specification for each of them are described in detail in \cite{myECCV2016}. 

\textbf{Landmark detection subnetwork.}  The architecture of the landmark detection network is based on a ResNet-152 model \cite{he2015deep}. The network was adapted for landmark localization by: (1) removing the fully connected layer alongside the preceding average pooling layer, (2) changing the stride of the 5th block from 2 to 1 pixels, and (3) adding at the end a deconvolution \cite{zeiler2011adaptive} followed by a fully convolutional layer. These changes convert the model to a fully convolutional network, recovering to some extent the lost spatial resolution. The ground truth was encoded as a set of N binary maps (one for each landmark), where the values located within the radius of the provided ground truth landmark are set to 1, while the rest to 0. The radius defining the ``correct location'' was empirically set to 7 pixels, for a face with a bounding box height equal to approximately 220 pixels. The network was trained using the pixel wise sigmoid cross entropy loss function.

\textbf{Landmark regression subnetwork.} As in \cite{myECCV2016}, the regression subnetwork plays the role of a graphical model aiming to refine the initial prediction of the landmark detection network. It is based on a modified version of the ``hourglass network'' \cite{newell2016stacked}. The hourglass network starts from the idea presented in \cite{long2015fully}, improving a few important concepts: (1) it updates the model using residual learning  \cite{he2015deep,he2016identity}, and (2) introduces an efficient way to analyze and recombine features at different resolutions. Finally, as in \cite{myECCV2016}, we replaced the original nearest neighbour upsampling of \cite{newell2016stacked} by learnable deconvolutional layers, and added another deconvolutional layer in the end that brings the output to the input resolution. The network was then trained using a pixel wise L2 loss function. 
 
 \begin{figure}
	\centering 
    \hspace{-1em}
    \vspace{-1em}
	\includegraphics[scale=0.3]{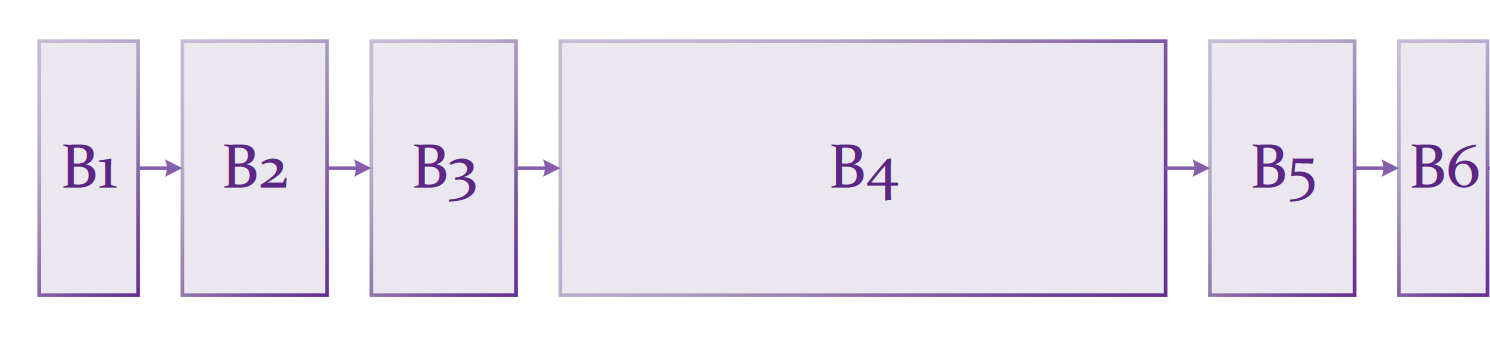}
	\caption{The architecture of the Z regression subnetwork. The network is based on ResNet-200 (with preactivation) and its composing blocks. The blocks B1-B6 are defined in Table \ref{table:DetContent}. See also text.}.
	\label{fig:Our3DNETZ}
\end{figure} 

 \begin{table}
	\begin{center}
		\caption{Block specification for the Z regression network. Torch notations (channels, kernel, stride) and (kernel, stride) are used to define the conv and pooling layers. The bottleneck modules are defined as in \cite{he2016identity}.}
        \scriptsize
		\label{table:DetContent}
		\begin{tabular}{| *6{>{\raggedright\arraybackslash}p{1.7cm}|}}
        \hline
			B1 & B2 & B3 & B4 & B5 & B6    \\ \hline
		 1x conv layer (64,7x7,2x2) 1x pooling (3x3, 2x2)   & 3x bottleneck modules [(64,1x1), (64,3x3), (256,1x1)]   & 24x bottleneck modules [(128,1x1), (128,3x3), (512,1x1)]  & 38x bottleneck modules [(256,1x1), (256,3x3), (1024,1x1)] & 3x bottleneck modules [(512,1x1), (512,3x3), (2048,1x1)] &  1x fully connected layer (66)  \\ \hline
		\end{tabular}
	\end{center}
\end{table}
\setlength{\tabcolsep}{1.4pt}

\subsection{Z regression}\label{ssec:method1d}
In this section, we introduce a third subnetwork for estimating the Z coordinate i.e. the depth of each landmark. As with the X,Y coordinates, the estimation of the Z coordinate is performed jointly for all landmarks. The input to the Z regressor subnetwork is the stacked regression heatmaps produced by the regression subnetwork alongside the input RGB image. The use of the stacked heatmaps is a key feature of the subnetwork as they provide pose-related information (encoded by the X,Y location of all the landmarks) and guide the network where to ``look'', explicitly showing where the depth should be estimated.

We encode each landmark as a heatmap using a 2D Gaussian with std=6 pixels centered at the X,Y coordinates of that landmark. The proposed Z regression network is based on the latest ResNet-200 network with preactivation modules \cite{he2016identity} modified as follows: in order to adapt the model for 1D regression, we replaced the last fully connected layer (used for classification) with one that has N output channels, one for each landmark. Additionally, the first convolutional layer of the network was modified to accommodate 3+N input channels. The network is described in detail in \figref{fig:Our3DNETZ} and Table \ref{table:DetContent}. All newly introduced filters were initialized randomly using a Gaussian distribution. Finally, the network was trained using the L2 loss:
\begin{equation}
l_2 = \dfrac{1}{N} \sum\limits_{n=1}^{N}  (\widetilde{z}_{n}-z_n)^2, 
\end{equation}
where $ \widetilde{z}_{n}$ and $ z_n$ are the predicted and  ground truth Z values (in pixels) for the $n$th landmark.

\subsection{Training}\label{ssec:methodtrain}
For training, all images were cropped around an extended (by 20-25\%) bounding box, and then resized so that the final cropped image had a resolution of 384x384 pixels. While batch normalization is known to prevent overfitting to some extent, we additionally augmented the data with a set of image transformations applied randomly: flipping, in-plane rotation (between $-35^o$ and $35^o$), scaling (between 0.85 and 1.15) and colour jittering. While the entire system, shown in \figurename{\ref{fig:Our3DNET}}, can be trained jointly from the beginning, in order to accelerate convergence we trained each task independently.  

For 2D (X,Y) landmark heatmap regression, the network was fine-tuned from a pretrained model on the large ImageNet \cite{deng2009imagenet} dataset, with the newly introduced layers initialized with zeros. The detection component was then trained for 30 epochs with a learning rate progressively decreasing from $ 1e-3 $ to $ 2.5e-5 $. The regression subnetwork, based on the "hourglass" architecture was trained for 30 epochs using a learning rate that varied from $ 1e-4 $ to $ 2.5e-5 $. During this, the learning rate for the detection subnetwork was frozen.  All the newly introduced deconvolutional layers were initialised using the bilinear upsampling filters. Finally, the subnetworks were trained jointly for 30 more epochs.

For Z regression, we again fine-tuned from a model previously trained on ImageNet\cite{deng2009imagenet}; this time we used a ResNet-200 network\cite{he2016identity}. The newly introduced filters in the first convolutional layer were initialized from a Gaussian distribution with std=0.01. The same applied for the fully connected layer added at the end of the network. During training, we used as input to the Z regression network both the heatmaps generated from the ground truth landmark locations and the ones estimated by the first task. We trained the subnetwork for about 100 epochs with a learning rate varying from $ 1e-2 $ to $ 2.5e-4 $. 

The network was implemented and trained using Torch7 \cite{collobert2011torch7} on two Titan X 12Gb GPUs using a batch of 8 and 16 images for X,Y landmark heatmap regression and Z regression, respectively.

\section{Experimental results}\label{sec:results}

In this section, we present the performance of our system on the 3DFAW dataset.

\textbf{Dataset.} We trained and tested our model on the 3D Face Alignment in the Wild (3DFAW) dataset. \cite{gross2010multi,yin2008high,zhang2014bp4d,jeni2016dense} The dataset contains images from a wide range of conditions, captured in both controlled and in-the-wild settings. The dataset includes images from  MultiPIE \cite{gross2010multi} and BP4D \cite{zhang2014bp4d} as well as images collected from the Internet. All images were annotated in a consistent way with 66 3D fiducial points. The final model was trained on the training set (13672 images)  and the validation set (4725 images), and tested on the test set containing 4912 images. We also report results on the validation set with a model trained on the training set, only.

\textbf{Metrics.} Evaluation was performed using two different metrics: Ground Truth Error(GTE) and Cross View Ground Truth Consistency Error(CVGTCE). 

GTE measures the average point-to-point Euclidean error normalized by the inter-ocular distance, as in \cite{sagonas2013300}. GTE is calculated as follows:

\begin{equation}
E(\mathbf{X},\mathbf{Y}) = \dfrac{1}{N} \sum\limits_{n=1}^{N}\dfrac{\norm{\mathbf{x}_n-\mathbf{y}_n}_2}{d_i}, 
\end{equation}
where $ \mathbf{X} $ is the predicted set of points, $ \mathbf{Y} $ is their corresponding ground truth and $ d_i $ denotes the interocular distance for the $ i $th image.

CVGTCE evaluates the cross-view consistency of the predicted landmarks of the same subject and is defined as follows:

\begin{equation}
E_{vc}(\mathbf{X},\mathbf{Y},T) = \dfrac{1}{N} \sum\limits_{n=1}^{N}\dfrac{\norm{s\mathbf{R}\mathbf{x}_n-\mathbf{y}_n}_2}{d_i}, 
\end{equation}
where $ P =\{s, \mathbf{R}, \mathbf{t}\} $ denotes scale, rotation and translation, respectively. CVGTCE is computed as follows: 

\begin{equation}
\{s,\mathbf{R},\mathbf{t}\} = \argmin_{s,\mathbf{R},\mathbf{t}}\sum\limits_{n=1}^{N}\norm{\textbf{y}_k-(s\mathbf{R}\mathbf{x}_k+\mathbf{t})}_{2}^{2}.
\end{equation}

\textbf{Results.} Table \ref{table:ResultsSeparateXYZ} shows the performance of our system on the 3DFAW test set, as provided by the 3DFAW Challenge team. As it can be observed, our system outperforms the second best method (the result is taken from the 3DFAW Challenge website) in terms of GTE by more than 22\%.

\begin{table}
	\begin{center}
		\caption{Performance on the 3DFAW test set.}
		\label{table:ResultsSeparateXYZ}
		\begin{tabular}{| *3{>{\raggedright\arraybackslash}p{1.8cm}|}}
			\hline
			Method & GTE (\%) & CVGTCE (\%) \\ \hline
		Ours & \textbf{4.5623} & \textbf{3.4767} \\ \hline
		Second best & 5.8835 & 3.9700 \\ \hline
		\end{tabular}
	\end{center}
\end{table}

\begin{figure*}[h]
	\centering
	\subcaptionbox*{a) xy.}{\includegraphics[height=1.6in]{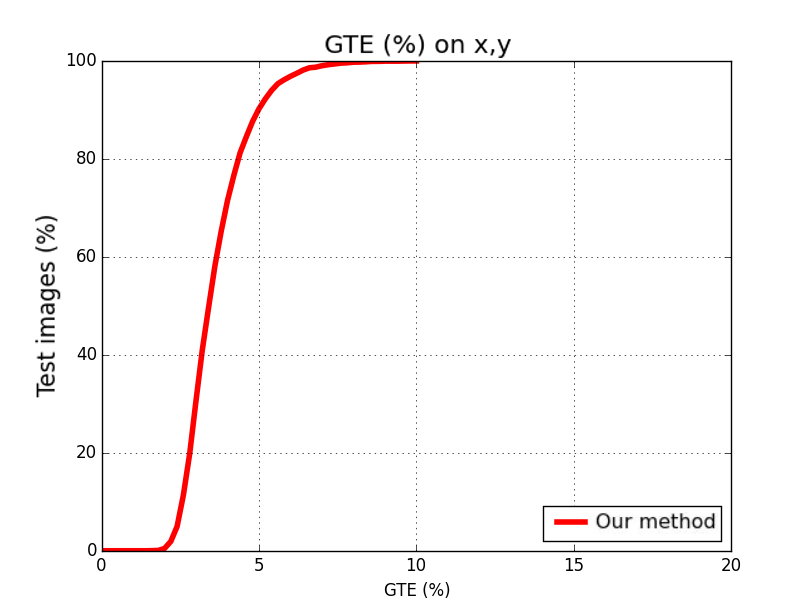}}
	\subcaptionbox*{b) xyz.}{\includegraphics[height=1.6in]{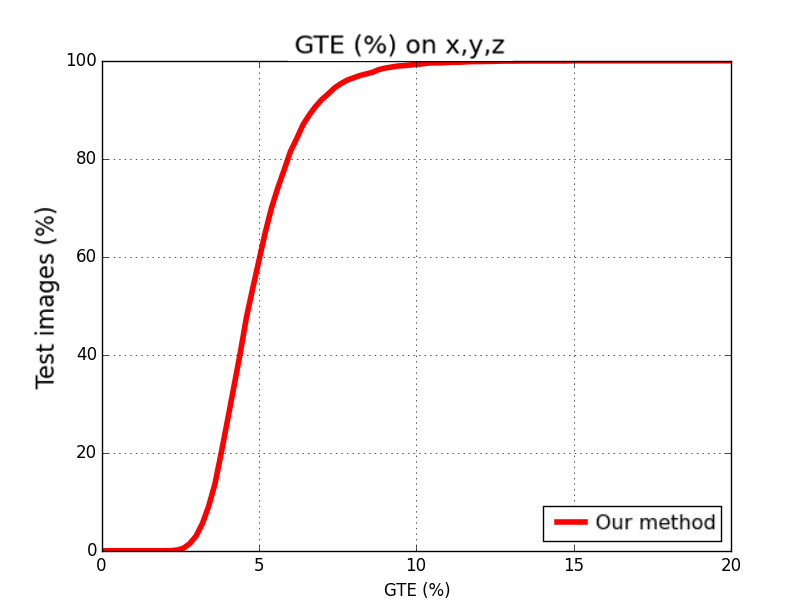}}
  \subcaptionbox*{c) x.}{\includegraphics[height=1.1in]{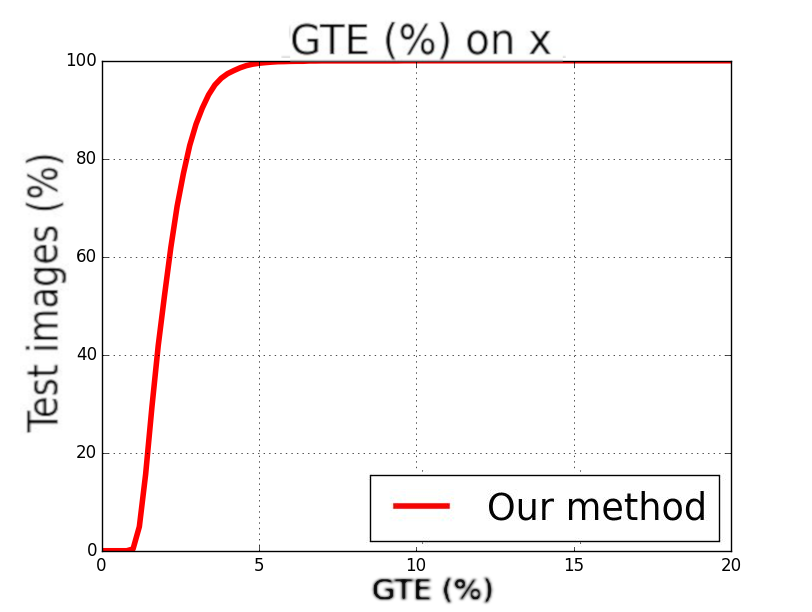}}
	\subcaptionbox*{d) y.}{\includegraphics[height=1.1in]{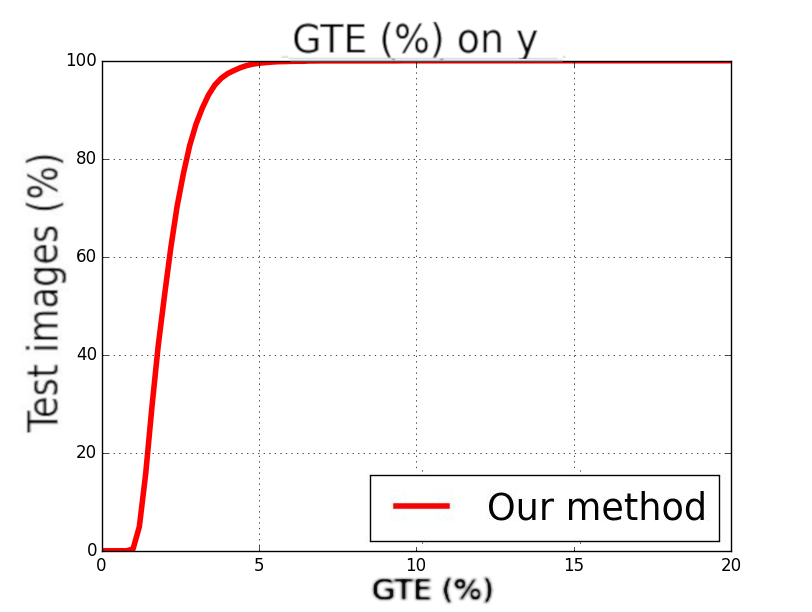}}
	\subcaptionbox*{e) z.}{\includegraphics[height=1.1in]{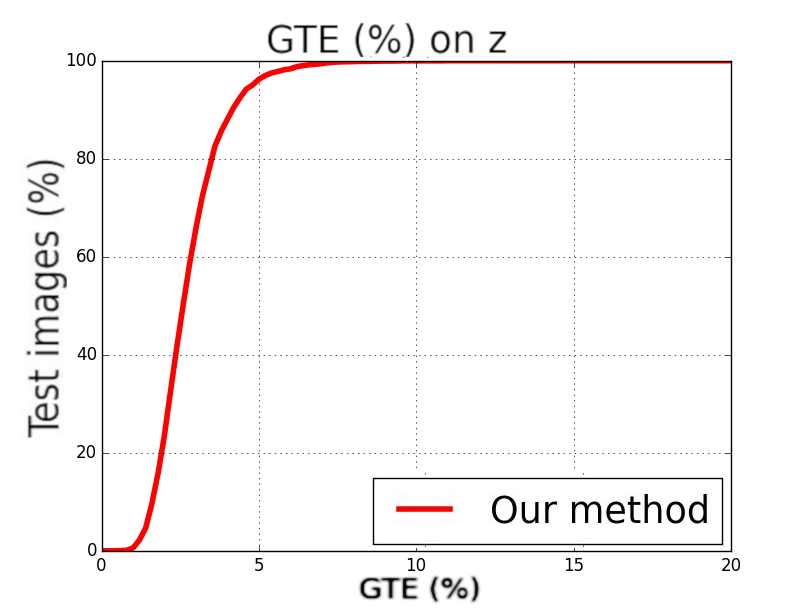}}
	\vspace{0.5em}
	\caption{GTE vs fraction of test images on the 3DFAW validation set, on (X,Y), (X,Y,Z), X alone, Y alone and Z alone.}
	\label{fig:CompCatsDogs}
\end{figure*}

\begin{figure}
	\centering 
	\includegraphics[scale=0.6, clip, trim=0cm 4.5cm 0cm 0cm,]{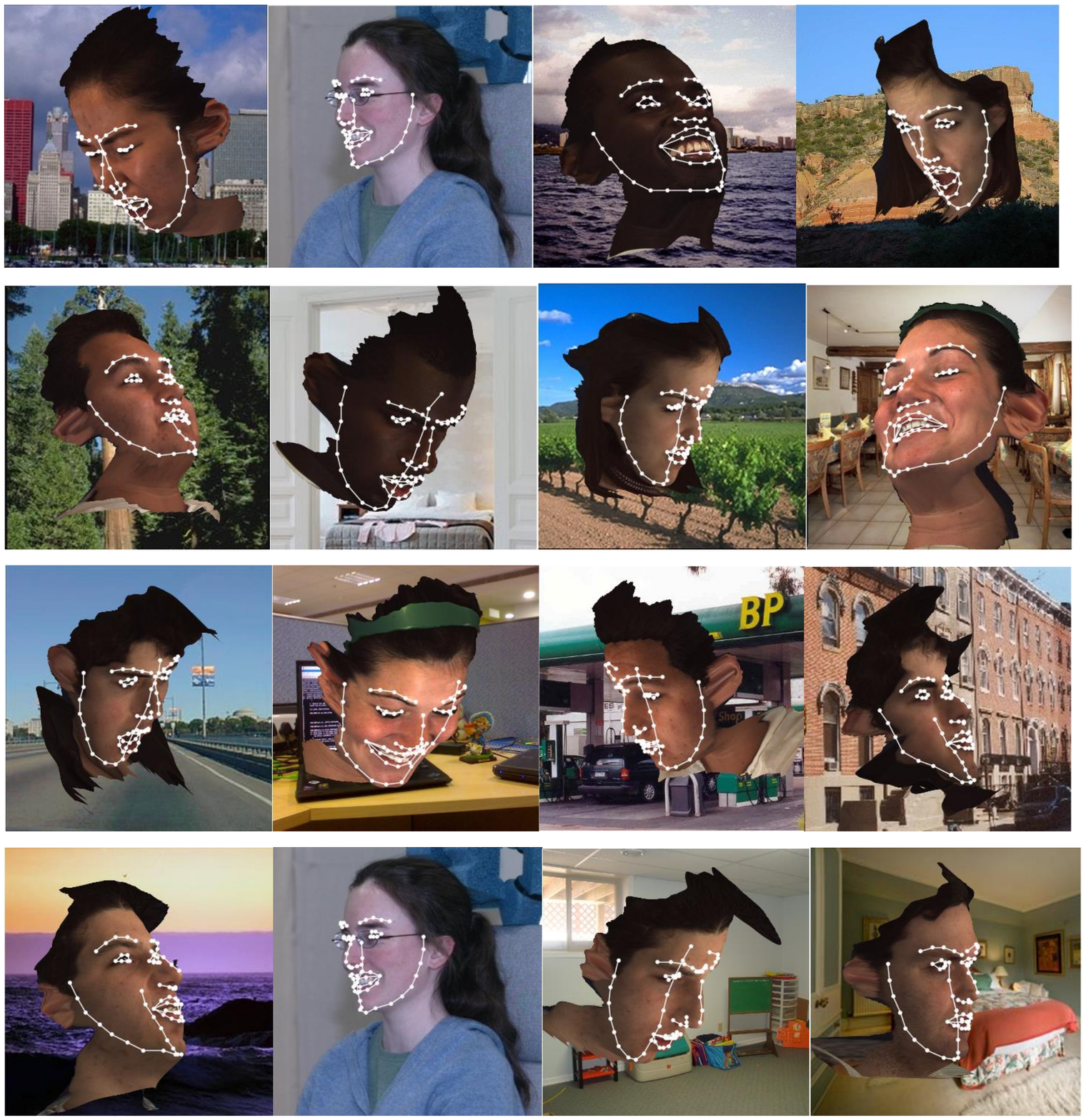}
	\caption{Fitting results produced by our system on the 3DFAW test set. Observe that our method copes well with a large variety of poses and facial expressions on both controlled and in-the-wild images. Best viewed in colour.}.
	\label{fig:3DExamples}
\end{figure}

In order to better understand the performance of our system, we also report results on the validation set using a model trained on the training set, only. To measure performance, we used the Ground Truth Error measured on (X,Y), (X,Y,Z), X alone, Y alone and Z alone, and report the cumulative curve corresponding to the fraction of images for which the error was less than a specific value. Results are reported in \figurename{\ref{fig:CompCatsDogs}} and Table \ref{table:axes}. It can be observed that our system performs  better on predicting the X and Y coordinates (compared to Z), but this difference is quite small and, to some extent, expected as Z is estimated at a later stage of the cascade. Finally, \figurename{\ref{fig:3DExamples}} shows a few fitting results produced by our system.

\section{Conclusions}\label{sec:conclusions}
In this paper, we proposed a two-stage CNN cascade for 3D face alignment. The method is based on the idea of splitting the 3D alignment task into two separate subtasks: 2D landmark estimation and 1D (depth) estimation, where the first one guides the second. Our system secured the first place in the 1st 3D Face Alignment in the Wild (3DFAW) Challenge.

\begin{table}[H]
	\begin{center}
		\caption{Performance on the 3DFAW validation set, on (X,Y), (X,Y,Z), X alone, Y alone and Z alone.}
		\label{table:axes}
		\begin{tabular}{| *2{>{\raggedright\arraybackslash}p{2.2cm}|}}
			\hline
			Axes & GTE (\%) \\ \hline
			xy & 3.6263 \\ \hline
            xyz & 4.9408 \\ \hline
            x & 2.12 \\ \hline
			y & 2.48 \\ \hline
			z & 2.77 \\ \hline
		\end{tabular}
	\end{center}
\end{table}

\section{Acknowledgment}

This work was supported in part by the EPSRC project EP/M02153X/1 Facial Deformable Models of Animals.

\clearpage

\bibliographystyle{splncs}
\bibliography{egbib}
\end{document}